\documentclass[letterpaper, 10 pt, conference]{ieeeconf} % Comment this line out if you need a4paper

\IEEEoverridecommandlockouts                              % This command is only needed if 
\usepackage{graphics} % for pdf, bitmapped graphics files
\usepackage{epsfig} % for postscript graphics files
\usepackage{mathptmx} % assumes new font selection scheme installed
\usepackage{times} % assumes new font selection scheme installed
\usepackage{amsmath} % assumes amsmath package installed
\usepackage{amssymb}  % assumes amsmath package installed
\usepackage{color} % Change font color

\usepackage{caption}
\usepackage{subcaption}

\usepackage{url}

\title{\LARGE \bf
Cross-view and Cross-domain Underwater Localization based on Optical Aerial and Acoustic Underwater Images
}

\author{}

\author{Matheus M. Dos Santos$^{1}$, Giovanni G. De Giacomo$^{1}$, Paulo L. J. Drews-Jr$^{1}$, Silvia S. C. Botelho$^{1}$% <-this % stops a spaced
\thanks{*This study was partly supported by CNPq and the Coordenacao de
Aperfeiçoamento de Pessoal de Nivel Superior - Brasil (CAPES) -
Finance Code 001. This paper is also a contribution of the INCT-Mar COI funded by CNPq Grant Number
610012/2011-8.}% <-this % stops a space
\thanks{$^{1}$Intelligent Robotics and Automation Group - NAUTEC,
Center for Computational Science - C3
Universidade Federal do Rio Grande - FURG,
Rio Grande, Brazil. E-mail:
{\tt\footnotesize \{matheusmachado,ggiacomo, paulodrews, silviacb\}@furg.br}\,.}%
\thanks{*This work has been submitted to the IEEE for possible publication. Copyright may be transferred without notice, after which this version may no longer be accessible.}%
}

\begin{document}

\maketitle
\thispagestyle{empty}
\pagestyle{empty}

%%%%%%%%%%%%%%%%%%%%%%%%%%%%%%%%%%%%%%%%%%%%%%%%%%%%%%%%%%%%%%%%%%%%%%%%%%%%%%%%   

\begin{abstract}

Cross-view image matches have been widely explored on terrestrial image localization using aerial images from drones or satellites. This study expands the cross-view image match idea and proposes a cross-domain and cross-view localization framework. The method identifies the correlation between color aerial images and underwater acoustic images to improve the localization of underwater vehicles that travel in partially structured environments such as harbors and marinas. The approach is validated on a real dataset acquired by an underwater vehicle in a marina. The results show an improvement in the localization when compared to the dead reckoning of the vehicle.

\end{abstract}

%\begin{IEEEkeywords} copy from RAL template
% Underwater Localisation, Acoustic Image, Neural Network
%\end{IEEEkeywords}

%%%%%%%%%%%%%%%%%%%%%%%%%%%%%%%%%%%%%%%%%%%%%%%%%%%%%%%%%%%%%%%%%%%%%%%%%%%%%%%%

\section{INTRODUCTION} \label{sec:introduction}

In autonomous vehicles, some localization methods go beyond a single view perception \cite{WU_2019}. \textit{Cross-view} localization methods combine data from different perspectives such as aerial and terrestrial images to estimate the terrestrial localization \cite{ Wolff_2016, Hu_2018_CVPR}. Typically, these methods localize street view images by matching georeferenced aerial images from satellites or drones.

\begin{figure}
    \centering
    \includegraphics[width=0.48\textwidth]{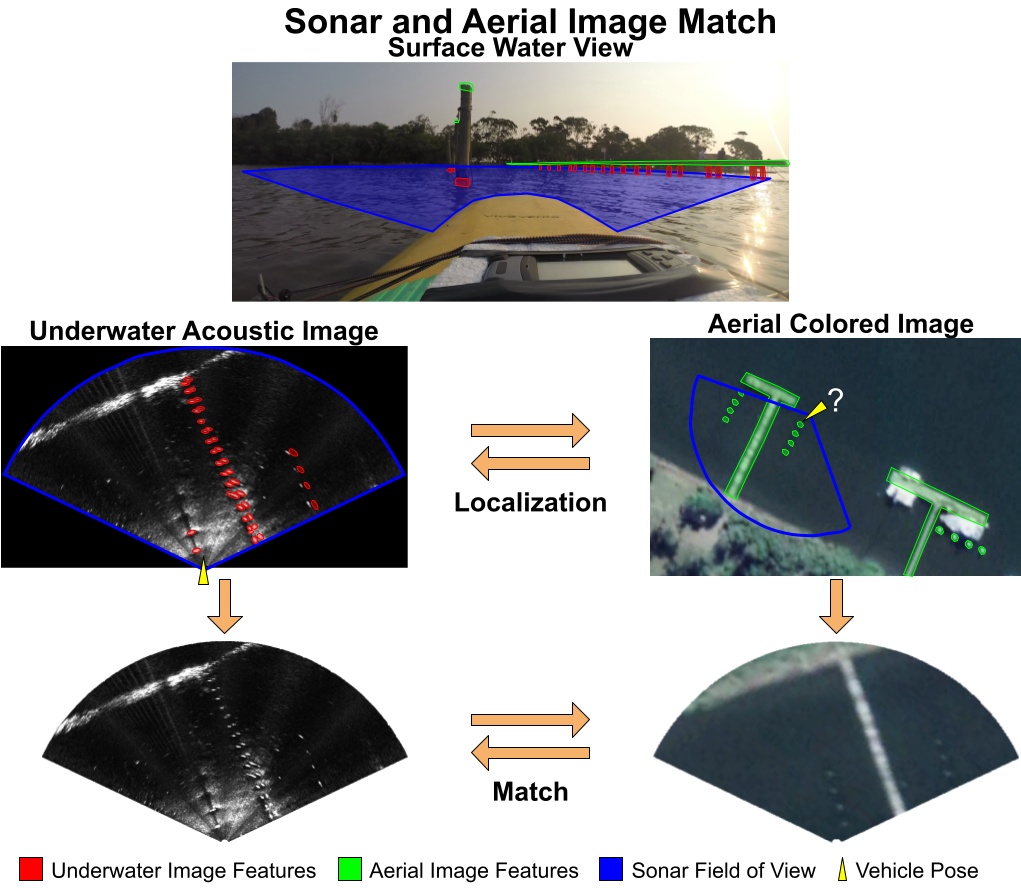}
    \caption{
    An underwater vehicle doted on multi-beam forward-looking sonar travels in a semi-structured marina environment. The localization is estimated using structures such as the pier and the shoreline, both visible on the range of the acoustic images. However, the vehicle gets lost when it moves to an open area where no structures are available. Aerial images of the environment can improve the localization and re-localize the vehicle when it returns to the shoreline region. Features highlighted in green and red are adopted in a \textit{Cross-view} and \textit{Cross-domain} matching system. A particle filter framework fuses both images and estimates the vehicle location using a Deep Neural Network (DNN) as an observation model. The system belief is built on aerial images.
    }
    \label{fig:intro}
\end{figure}

In this study, we localize an underwater vehicle by matching its underwater acoustic images with aerial georeferenced satellite images. It configures a \textit{Cross-View} localization problem because the underwater acoustic images provide a frontal view and the aerial images provide a top view of the scene. In addition, we have the \textit{Cross-Domain} problem because of the acoustic and the optical domains of the images. Fig. \ref{fig:intro} represents the \textit{cross-view} and \textit{cross-domain} localization problem addressed in this work. While the gray-scale underwater acoustic images only provide distances and shapes of the observed objects the aerial optical images provide rich texture and color.

Our method is designed to operate in partially structured environments such as marinas and harbors. These places provide stable features such as piers, stones, and the shoreline that can be observed in both aerial and underwater images.

The problem of matching of underwater acoustic and optical aerial images was previously addressed on \cite{Santos19, Machado_2020,Giacomo_2020}. Santos \textit{et al.} \cite{Santos19, Machado_2020} proposed a Deep Neural Network (DNN) based on a Siamese architecture \cite{chopra_2005} that handles the cross-domain problem by training two independent networks. Giacomo \textit{et al.} \cite{Giacomo_2020} presented an approach inspired by Generative Adversarial Network (GAN) \cite{Goodfellow_2014} and Triplets Network \cite{Hoffer_2015}. They trained two models using a quadruplet strategy, an adaptation of the triplet strategy with an additional anchor image. The paper showed the latter approach achieved better results and performance. However, none of the previous work addressed the underwater localization problem, only the matching problem.
 
The main contribution of this work is a new underwater localization framework based on a map built from georeferenced aerial image and data association with acoustic images on the Adaptive Monte Carlo Localization (AMCL) algorithm \cite{Thrun_2006}. The method allows the robot localization in a GPS-denied environment such as the water.
 
The Monte Carlo Localization, also known as Particle Filter, is a well-known localization method that can model a multi-modal non-Gaussian probabilistic distribution function by spreading hypotheses in the map known as particles. The particles allow us to select the most likely regions on satellite images (map) that match the acoustic underwater image (perception).
 
The method is suitable for data association because we crop the aerial image such as both images, acoustic and aerial, have the same size and shape. Aerial images cover a large area and have a larger size than underwater acoustic images. Therefore, the aerial image must be cropped before its use on the image matching system.

The proposed localization framework can be applied to unmanned underwater vehicles or hybrid aerial-underwater vehicles \cite{Drews_2014,Neto2015,mercado2019,Horn_2020} to perform tasks, such as inspection and surveillance on harbors and marinas. This work is validated on real data collected by an underwater vehicle in a marina and optical aerial images acquired from a satellite. 
 
The results showed the method can localize the underwater vehicle using satellite images and achieving better results than dead reckoning.
 
This paper is organized as follows: Section \ref{sec:related_works} presents related work,  Section \ref{sec:methodology} explains our proposed pipeline and each step,  Section \ref{sec:results} shows the experimental results with a real dataset, and finally, Section \ref{sec:conclusion} summarizes our contributions and outlines our future work.

\section{RELATED WORK} \label{sec:related_works}

Methods to fuse cross-view and cross-domain data on aerial and underwater domains to estimate underwater localization are not widely explored in the literature, being the first to the best of our knowledge. One of the main issues of this kind of method is the cross-view and cross-domain matching. Some work propose a cross-view matching of terrestrial and aerial images to geolocalize terrestrial images. Xiang Gao \textit{et al.} \cite{GAO_2018} presents a review of these methods and classifies them as image-based and structure-based methods. Most of the aerial and terrestrial image matching methods explore self-similar features, semantic features, or Deep Neural Networks approaches. 

Self-similar feature methods \cite{Bansal_2016, Wolff_2016} look for features present on structures such as skyscrapers facades that can be detected in both views. For example, street view and aerial images are matched using shared features such as texture and color from the same skyscrapers facade. 

Methods based on structures \cite{Li_2014, Majdik_2013, Vis_2014} look for the same structures in both aerial and terrestrial view. Typically, this approach is applied when the environment does not provide enough features to compare the images. Semantic-based methods combine external information about the observed scene to perform matching \cite{Lin_2013_CVPR,Castaldo_2015}. 

The deep learning methods perform matching based on learned features from the image datasets. Lin \textit{et al.} proposed the first deep learning network called Where-CNN \cite{Lin_2015_CVPR} to learn feature embedding for image matching. Workman \textit{et al.} \cite{Workman_2015_CVPR,Workman_2015_ICCV} obtained state-of-the-art results on wide-area image geolocation with a new cross-view training strategy for learning a joint semantic feature representation for aerial images. Tian \textit{et al.} \cite{Tian_2017_CVPR} proposed a matching approach of terrestrial and aerial images based on urban buildings. 

  Karkus \textit{et al. }\cite{Karkus_2018_ParticleNet} propose an end-to-end Recurrent Neural Network (RNN) that encodes a full differential particle filter algorithm. The method is designed to localize an indoor vehicle with a single camera. The simulated dataset House3D is adopted to train and validate the method. It is an interesting work because it allows training the network without labeled data. However, the authors do not test the system on a real-world environment, and it places in check if an end-to-end approach would be able to localize on real environment data.

  Localization of underwater acoustic images using aerial satellite images has similar aspects with the terrestrial image localization with aerial images. Both problems involve cross-view matching. However, only the acoustic data is obtained in a diverse domain. Furthermore, the GPS system is usually available in ground robots while the underwater environment is GPS-denied.
  
  Giacomo \textit{et al. } proposed in \cite{Giacomo_2021} a neural network that translates a gray-scale acoustic image into a colored aerial satellite image. The proposed network is based on the U-Net architecture and employs techniques such as dilated convolutions, guided filters, and the Deep Convolutional Generative Adversarial Network (DCGAN).
  
  A few work study the underwater acoustic and aerial image match. Santos \textit{et al.} \cite{Machado_2020} proposed a Deep Neural Network (DNN) based on a Siamese architecture \cite{chopra_2005} that handles the cross-domain problem by training two independent networks.
  
  Giacomo \textit{et al. }\cite{Giacomo_2020} proposed a cooperative approach for training multiple networks to reduce the image dimensionality belonging to the same domain of the present work. Their method consists of cooperatively optimizing two neural networks that share the same architecture but not the same weights. Then, these networks update their weights following the triplet objective function. In the end, it is possible to use the trained networks to extract vectors that encode the images fed into them. Afterward, a distance between the extracted vectors can be calculated, such as the Euclidean distance. This network makes possible tasks such as matching and ranking, which are of primordial importance to the present work.

  Santos \textit{et al.} \cite{Machado_2020} and Giacomo \textit{et al. }\cite{Giacomo_2020} research provide image matching methods but none of them estimate localization. In this work we adopt the quadruplet neural network from Giacomo \textit{et al. }\cite{Giacomo_2020} in a complete framework that estimates the vehicle localization.

  Structure-based approaches inspired our method because the structures can be easily identified on sonar images. Our approach uses the structure of sonar images for the localization in a map generated by the aerial image, in a probabilistic framework based on Adaptive Monte Carlo Localization (AMCL) \cite{Thrun_2006}.

\section{METHODOLOGY}\label{sec:methodology}
\begin{figure*}[htb]
    \centering
    \includegraphics[width=0.98\textwidth]{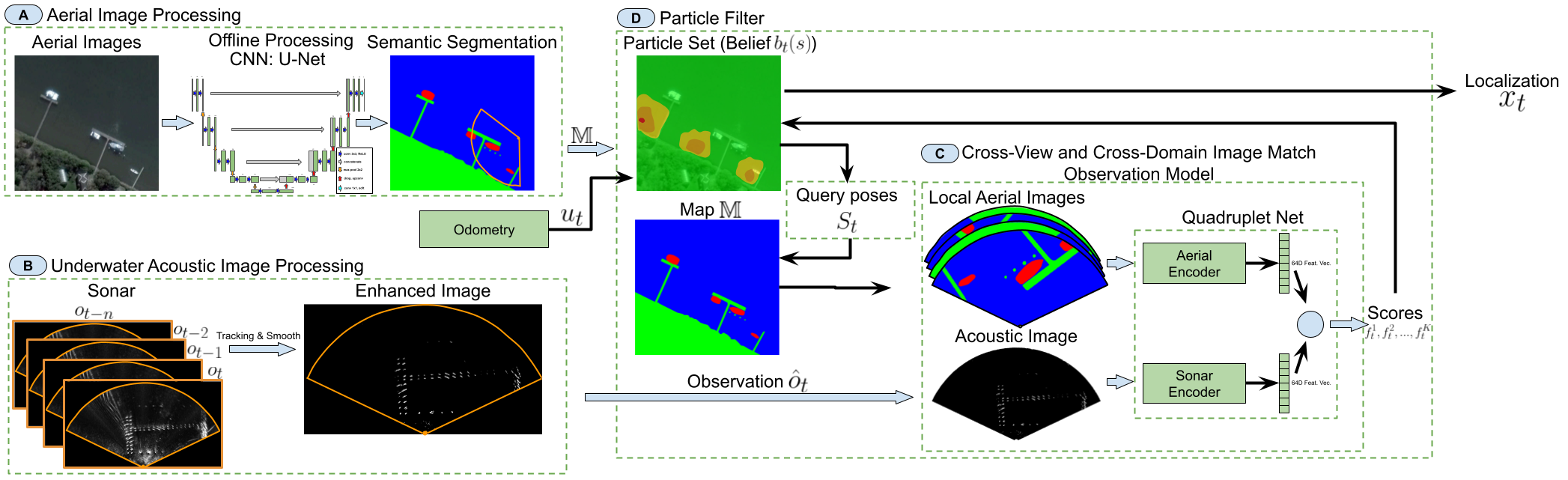}
    \caption{Proposed framework for cross-view (top and frontal) cross-domain (optical and acoustic) vehicle localization using acoustic and satellite images. Initially, the satellite image is semantically segmented in an offline process. The Underwater Acoustic Image Processing  performs image enhancement based on the alignment of a batch of acoustic images $O_t = \{ o_{t},o_{t-1},...,o_{t-n} \}$.
    The particle filter adopts the satellite image as the map $\mathbb{M}$, the enhanced acoustic images $\hat{o}_{t}$ as the observations and the vehicle odometry as control signal $u_t$ and estimates the underwater vehicle state $Y_t$.
    }
    \label{fig:pipeline}
\end{figure*}

 The underwater localization method is shown in Fig. \ref{fig:pipeline}. A particle filter algorithm \cite{Thrun_2006} estimates the vehicle localization $x_t$ based on the current observation $o_t$, vehicle control $u_t$, and the map of the scene $\mathbb{M}$. Where the observation $o_t$ is an underwater acoustic image and $\mathbb{M}$ is the aerial satellite image.
 
 The system has the following four main processes: A - Aerial Image Processing, B - Underwater Acoustic Image Processing, C - \textit{Cross-View} and \textit{Cross-Domain} Image Match, and D - Particle Filter. Step A runs once and offline. The remaining process runs in parallel in the computation graph of the Robot Operating System (ROS) \cite{ros}.  Each process is described in the following sections.

\subsection{Aerial Image Processing}

 The Aerial Image Processing employs the neural network U-Net to semantic segment the aerial image in three classes: stationary structures, movable objects, and the water highlighted in green, red, and blue on Fig. \ref{fig:pipeline}-A. The process is better explored and better described in \cite{Machado_2019}. Movable objects such as boats cannot be trusted on the localization problem. However, stationary structures such as piers and stones are relevant information to incorporate into the system.
 This process runs once and offline before the mission starts. Then, the segmented images are loaded into the vehicle memory as a map $\mathbb{M}$. Manual segmentation also can be employed since the process runs offline with no significant time restrictions.

\subsection{Underwater Acoustic Image Processing}

The Acoustic Image Processing smooths the current acoustic image referenced as observation $o_t$ by aligning a batch of previous images $o_{t-1},o_{t-2},...,o_{t-n}$. The alignment process starts by transforming the acoustic images into a 2D point cloud. First, a border detection based on image gradient segments the image. Each 2D point is defined as the centroid of the segments. The Iterative Closest Point (ICP) algorithm finds the affine transform between the 2D point clouds. Then, all images are transformed into the current image view point and smoothed by averaging the pixels.

\subsection{Cross-View and Cross-Domain Image Match} \label{sec:match}

The Image Match process uses the Quadruplet neural network proposed by Giacomo \textit{et al. }\cite{Giacomo_2020} running in a Graphics Processing Unit (GPU). The process evaluates the similarity between a batch of semantically segmented satellite images and the current acoustic image as represented on Fig. \ref{fig:pipeline}-C.

Two neural networks are used to encode acoustic and segmentation images. Each network is used for one image domain, \textit{i.e.}, one for acoustic, and another for a segmented and cropped aerial image. A cooperative approach is used to train the encoding networks. 

It is worth noting that both the acoustic and the segmented image encoder follow the same architecture. However, the weights are not shared, leading to different networks, as described in \cite{Giacomo_2020}. Due to the cooperative approach, the networks are trained to perform Metric Learning, by way of a triplet loss function. As a result, the networks produce vectors of reduced dimensionality from the original images. By taking the Euclidean distance between the produced vectors, a meaningful metric can be obtained. This metric is closer to zero when the images are similar and grows as they diverge. Therefore, the Euclidean distance between the vectors can be used to generate a rank of matching images.
 
This process receives a batch of satellite images and one acoustic image. The images are re-scaled to the lower resolution of $128\times256$ to fit on the Quadruplet Neural Network input. This process output is a normalized matching score $f_t$ of each satellite image where $t$ is the timestamp of the acoustic image.

A inversion on the Euclidean distance between the vectors are performed to generate a matching score $f_t$ considering all images of the batch. Defining $d_t^{k}$ as the distance between the satellite image $k$ and current acoustic image, we find $dmax_{t} = max(d_{t}^{1},d_{t}^{2},...,d_{t}^{K})$ and $dmin_{t} = min(d_{t}^{1},d_{t}^{2},...,d_{t}^{K})$ and applying $f_t^{k} = dmax_{t} - d_t^{k} + dmin$. The final scores are normalize such as $f_t^{k} = \frac{\hat{f}_t^{k}}{ \sum _{i=1}^{K} \left( \hat{f}_t^{k} \right) } $ where the $K$ denotes the number of particles in the current set, and $t$ is the timestamp.

\subsection{Particle Filter}

The particle filter algorithm estimate the vehicle state based on belief $b_t(s)$ that is approximated by a set of $K$ particles, $b_t(s) \approx \{(s_t^k, w_t^k)\}$ where $1 \leq k \leq K$, $w_t^k$ is the particle weight, $\sum_k w_k = 1$ and each particle $s_t^{[k]}$ represents an hypothesis of the vehicle pose in the world at time $t$. The belief is built on the map $\mathbb{M}$, \textit{i.e} the semantic segmented satellite image. The transition model $T$ and the observation model $Z$ update the particle set. 

\subsubsection{Transition Model} 

The transition model $T$ estimates the particle state $s_{t}^{k}$ based on previous state $s_{t-1}^{k}$ and the vehicle control signal $u_{t}$ such as $s_{t}^{k} \sim  T(s_t|u_t, s_{t-1}^k)$. A constant velocity model estimates the vehicle motion based on its control signal. The body-centered forward motion $v_x$ is converted to a motion on the world frame using current vehicle orientation measured by a compass. The transition model updates all particles every new control message $u_t$.

\subsubsection{Observation Model} 
 
\begin{figure}
    \centering
    \includegraphics[width=0.42\textwidth]{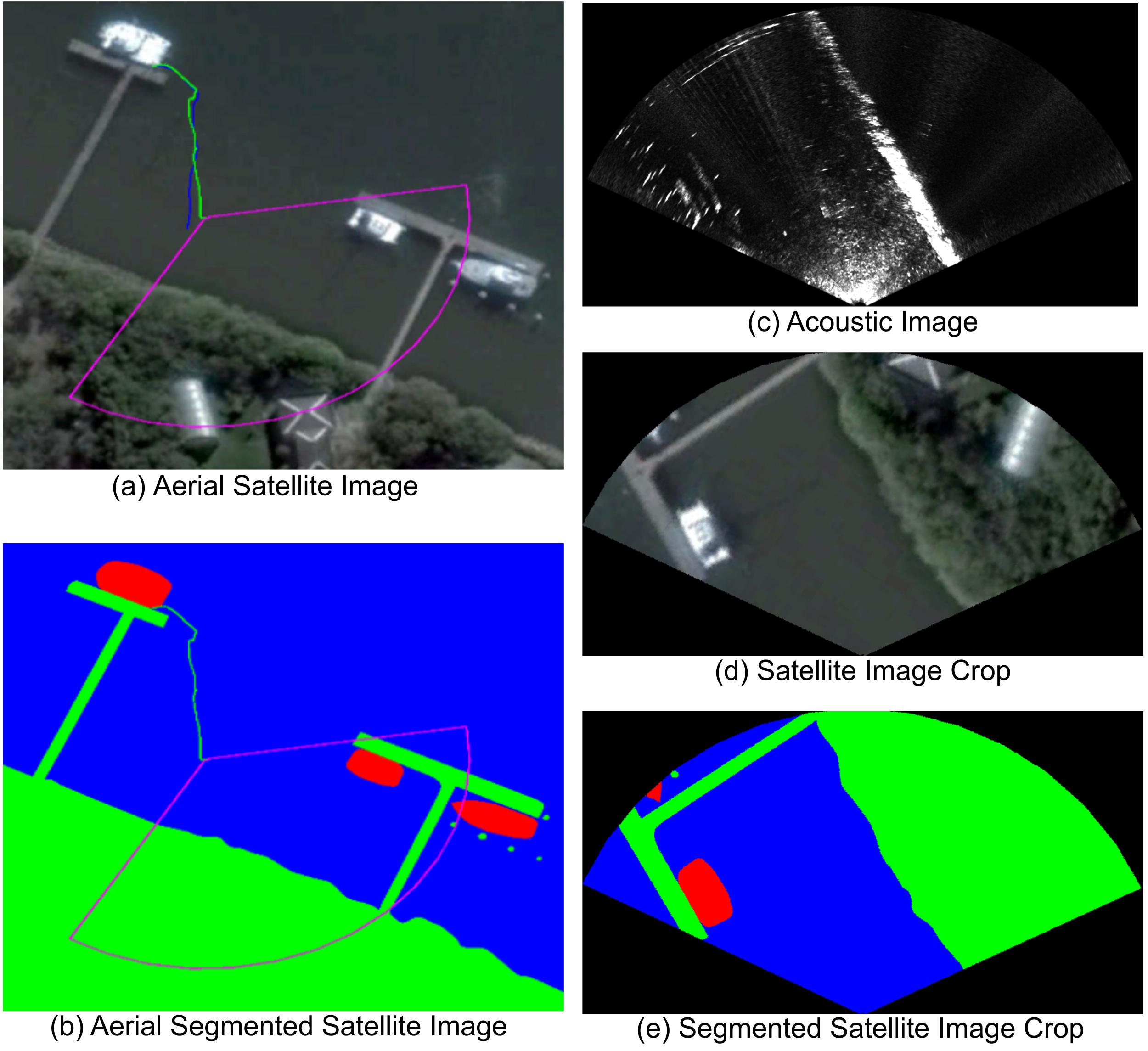}
    \caption
    { A particle state $s_{t}^{k}$ is converted into a crop of the semantically segmented satellite image. The crop has the same shape and scale as the acoustic images.}
    \label{fig:sat_crop}
\end{figure}

The observation model $Z$ incorporates an enhanced acoustic image $\hat{o}_t$ on the particle filter set. It computes the likelihood $f_{t}^{k}$ of $\hat{o}_t$ given the particle state $s_{t}^{k}$ and the map $\mathbb{M}$ such as $f_{t}^{k} = Z(\hat{o}_t| s_{t}^{k},\mathbb{M})$. This is a key point of the method because the particle state $s_{t}^{k}$ allows us to crop an aerial satellite image of any size in such a way that it has the same shape and scale of the acoustic images, as shown in Fig. \ref{fig:sat_crop}.
Each particle generates an aerial satellite image crop that is compared with the current acoustic image using the matching process described on Sec. \ref{sec:match}. As result, the normalized score $f_{t}^{k}$ of each particle $k$ updates the particle weight $w_{t}^{k}$. 

This process runs for each new underwater acoustic image. Initially, a information test is made, \textit{i.e.} when less than 2\% of the image pixels are non-zero, we consider the image as non-informative and discard it due to lack of information. Otherwise, the image feeds the observation model that updates the particle's weight. This verification is useful because the vehicle can be in open waters, where there is no structure to match the satellite image. In this case, the vehicle relies on the odometry. When it returns to a region with structures, the algorithm corrects the localization error using the satellite image.

\subsubsection{Initialization} 

The particle set is randomly initialized following a Gaussian distribution with mean $\mu$ on the last vehicle position before dive and a standard deviation $\sigma$.

\subsubsection{Re-sampling}

A re-sampling process is performed on the particle set using the Survival of the fittest principle \cite{Thrun_2006}, where the unlikely particles are replaced by the more likely ones. We adopted the roulette wheel approach which uses binary search and re-sample all particles in $O(M log M)$. After the re-sampling, we check for bad particles with the following tests:
  
\begin{itemize}
  \item Is the particle on open water? We check if the cropped image views stationary structures or not, \textit{i.e} green pixels on the semantic segmented aerial image.
  \item Is the particle laying on the floor? We check if the particle position is on a stationary structure, \textit{i.e} a green pixel on the semantically segmented satellite image.
  \item Is the particle out of the map borders? We check if the particle position is out of the satellite image boundaries.
\end{itemize}

In any positive case, the particle is re-sampled again using a higher standard deviation. As a last chance, when a particle is re-sampled more than ten times in the same iteration, a uniform random pose in the map is set.

\section{RESULTS}\label{sec:results}

 Our method is evaluated on real underwater scenario of the dataset ARACATI 2017 using Robot Operating System (ROS) \cite{Quigley_2009}.%, rosbag, plotjuggler and rqt\_gui tools.

\subsection{Dataset ARACATI 2017}

\begin{figure}
    \centering
    \includegraphics[width=0.9\linewidth]{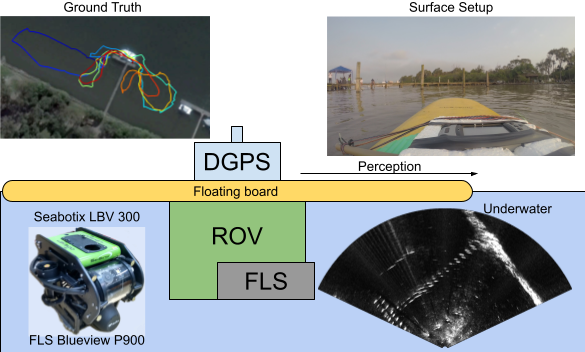}
    \caption{On the dataset ARACATI 2017, a floating board holds a DGPS on surface water and a Seabotix LBV 300 underwater ROV. It allows collecting precise position data as well as underwater acoustic images.}
    \label{fig:aracati_2017}
\end{figure}

The dataset ARACATI 2017 was recorded on the marina of Yacht Club of Rio Grande Brazil with a Seabotix Little Benthic Vehicle LBV 300-5 and a Forward-Looking Sonar (FLS) BlueView P900 \cite{Santos_2019}. The vehicle was attached below a floating board such as it stays underwater while a Differential Global Position System DGPS stays in the top of the board outside water as shown in Fig. \ref{fig:aracati_2017}.

We evaluate the performance of the system by collecting acoustic images, compass, odometry and DGPS as ground truth at a maximum speed of 0.6 m/s. The marina has a minimum depth of 1 meter and a maximum depth of 5 meters. The coastline is covered by stones that have a strong acoustic signature. Pontoon objects, moving boats, fish, and acoustical signatures from the seafloor and surface are present in the sonar data\footnote{Dataset ARACATI 2017 is available at \url{https://github.com/matheusbg8/aracati2017} }.

\subsection{Experimental results}

  The Localization and Matching process are evaluated on dataset ARACATI 2017 using a fixed number of 120 particles and the parameters of Table \ref{tab:parameters}. Where $\sigma$ is the standard deviation and $\mu$ is the mean of a Gaussian distribution function used to estimate the state $[x,y,\theta]^T$ of each particle $s_t^{k} \in S$. The initial position guess is the last GPS message before dive.
  
\begin{table}[]
\caption{Experiment Parameters}
\label{tab:parameters}
\begin{tabular}{l|c|c|}
\cline{2-3}
                                               & $\sigma$ & $\mu$                              \\ \hline
\multicolumn{1}{|l|}{Particle Re-sampling}  & $0.15$     & On Particle Selected with Roulette             \\ \hline
\multicolumn{1}{|l|}{Particle Initialization}     & $0.5$   & Initial Position Guess \\ \hline
\multicolumn{1}{|l|}{Bad Particle Re-sampling} & $15$     & On Particle Selected with Roulette \\ \hline
\end{tabular}
\end{table}

\begin{figure}
    \centering
    \includegraphics[width=0.9\linewidth]{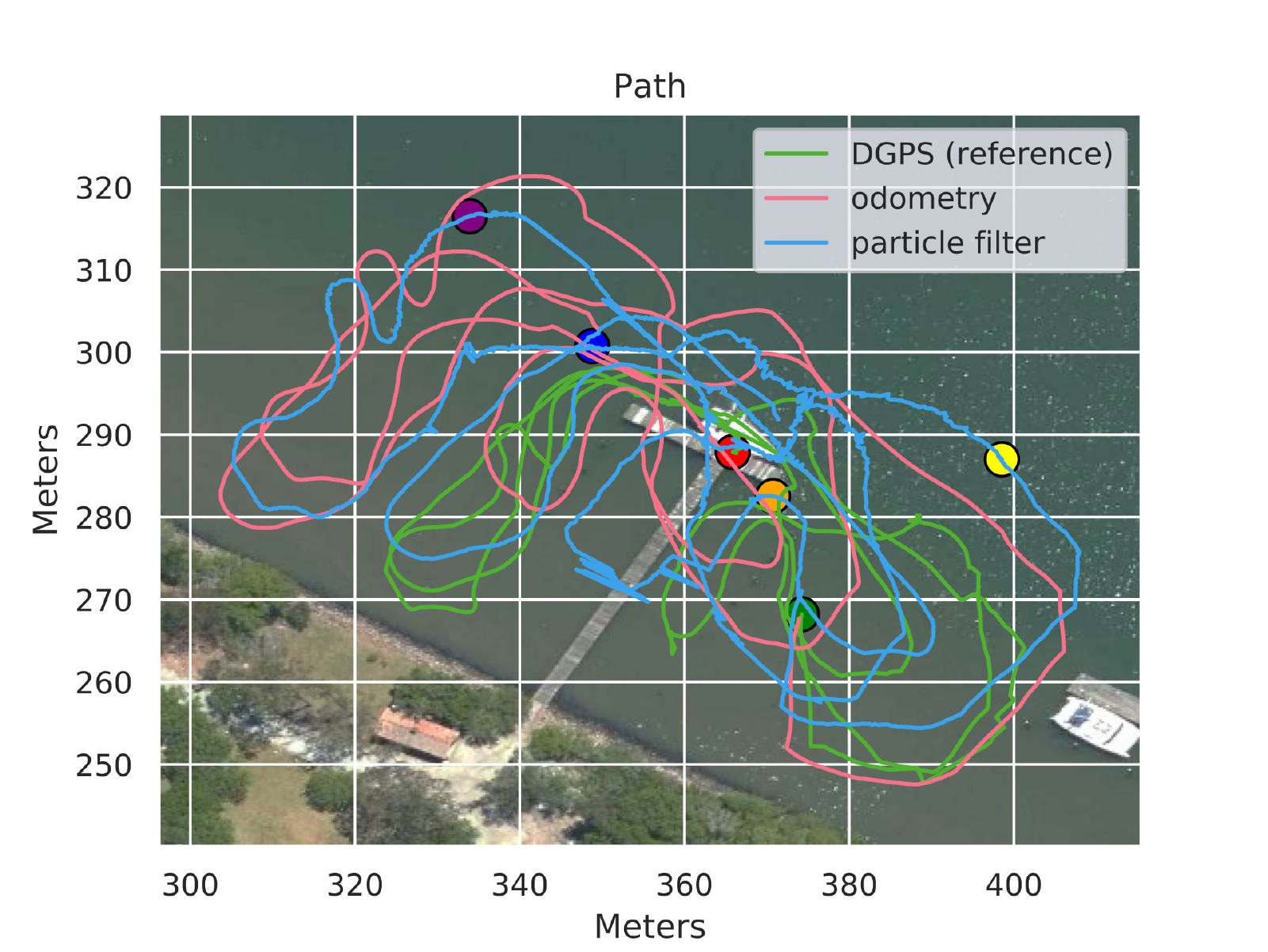}
    \caption{Vehicle path in meters estimated by dead reckoning in red, particle filter in blue, and DGPS reference in green. The colored circles highlight timestamps when features were not available on sonar or when the method starts to correct the localization based on cross-view and cross-domain matching.}
    \label{fig:path}
\end{figure}

\begin{figure}
    \centering
    \includegraphics[trim=0 3 0 0, clip,width=0.45\textwidth]{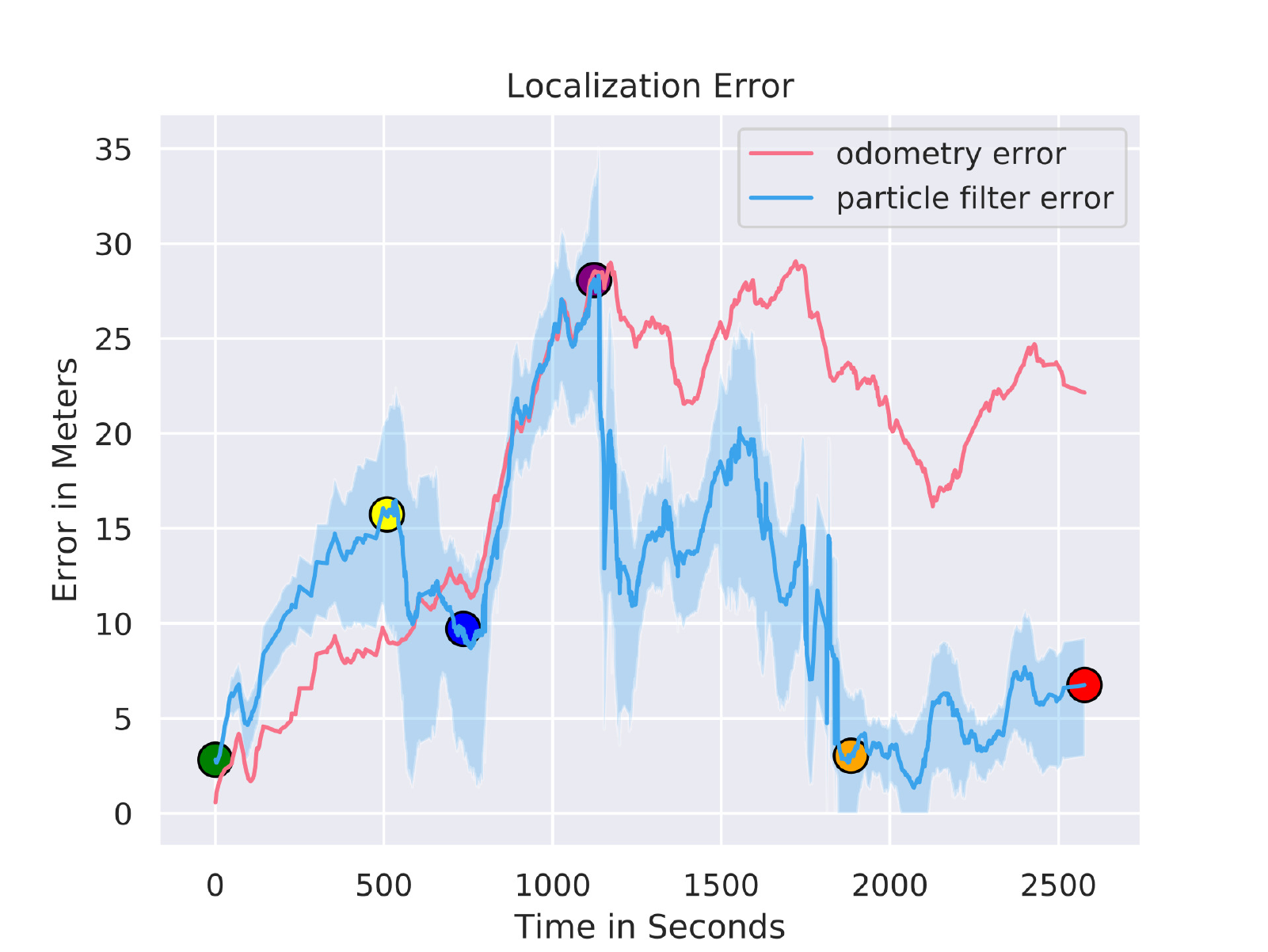}
    \caption{Time series in seconds with localization error in meters from dead reckoning in red and particle filter in blue. The light blue color indicates the uncertainty of the particle filter, and the six colored circles indicate the same timestamps of Figure \ref{fig:path}}
    \label{fig:error}
\end{figure}

 The experiment resulted on the path shown in Fig. \ref{fig:path}. The green line indicates the DGPS data adopted as ground truth, the blue line indicates the path from our localization method, and the red line indicates the dead reckoning. The localization error in meters of the dead reckoning and our method relative to the DGPS is shown in Fig. \ref{fig:error}.

\addtolength{\textheight}{-1cm}

The vehicle path crosses the six marks represented by the circles in figures \ref{fig:path} and \ref{fig:error}. It starts on the green circle, moves to yellow, blue, purple, orange, and ends on the red circle after 41 minutes. The results show that most of the time our method had a smaller localization error than the vehicle's dead reckoning. However, some aspects must be considered. Our system is based on images and depends on structures to perform cross-view and cross-domain matching. When features are not available, our method relies most on odometry.

In the first seconds of the experiment, the particles were initializing while the vehicle was moving. It led the particles to lose part of the initial control signals and transition model updates, resulting in a higher localization error. In the next ten minutes, the localization remains high because of the lack of features on the acoustic images. After the yellow mark, the sonar detected the central pier, and the localization error dropped while the vehicle was getting closer to the structures. Then the vehicle traveled to the left side of the pier and moved in an open area between blue and purple marks. At this point, the method relied most on odometry again because of the leak of acoustic features. After the purple mark, the vehicle returned to the central pier. Its localization got corrected thanks to our observation model and the particle filter. After this point, the method had a better localization estimation than the dead reckoning\footnote{The experimental results are available in video at \url{https://youtu.be/UOAcSODbaIw}}.

The experiment ran on a computer with a Processor Ryzen 7 2700x and a GPU NVIDIA RTX 3070. The image match evaluated 120 particles in 0.72 secs (1.39 Hz) and used 3.3 GB of RAM and 6 GB of VRAM. The particle filter process used 71 MB of RAM and the average execution time was 1.207 secs (0.83 Hz). The process runs in parallel, and their implementation is not optimized. The top speed of the underwater vehicle in the experiment was 0.63 m/s, and the results were achieved by processing one acoustic image every 4 seconds. We believe the system can run on an embedded platform such as NVIDIA Jetson AGX Xavier which has 32 GB of shared memory.

The current method does not estimate roll and pitch angles. We adopted an open frame vehicle calibrated to passively stabilize these angles. However, we believe that our method can be adapted to operate in parallel with the attitude control of a typical AUV.

\section{CONCLUSIONS}\label{sec:conclusion}

We proposed a new underwater vehicle localization method based on cross-view and cross-domain image match and validated it in a real experiment. The system uses a Deep Neural Network for image matching and an adapted Particle Filter for state estimation. The method can localize an underwater vehicle in a partially structured environment with acoustic images from a Forward-Looking Sonar and optical aerial images of the environment from satellites.

As future work, we want to compare different underwater localization approaches in a new experiment where we can improve the odometry model of the vehicle and test different observation models on the particle filter algorithm. We are also planning to perform real-time tests on embedded platforms such as NVIDIA Jetson Xavier running together with the robot using high-resolution images collected by Unmanned Aerial Vehicles (UAV).


%%%%%%%%%%%%%%%%%%%%%%%%%%%%%%%%%%%%%%%%%%%%%%%%%%%%%%%%%%%%%%%%%%%%%%%%%%%%%%%%

%%%%%%%%%%%%%%%%%%%%%%%%%%%%%%%%%%%%%%%%%%%%%%%%%%%%%%%%%%%%%%%%%%%%%%%%%%%%%%%%

%%%%%%%%%%%%%%%%%%%%%%%%%%%%%%%%%%%%%%%%%%%%%%%%%%%%%%%%%%%%%%%%%%%%%%%%%%%%%%%%
%\section*{APPENDIX}


%\section*{ACKNOWLEDGMENT}

%%%%%%%%%%%%%%%%%%%%%%%%%%%%%%%%%%%%%%%%%%%%%%%%%%%%%%%%%%%%%%%%%%%%%%%%%%%%%%%%

%References are important to the reader; therefore, each citation must be complete and correct. If at all possible, references should be commonly available publications.

%\bibliographystyle{IEEEtran}

%\bibliography{IEEEabrv,bib}

\begin{thebibliography}{10}
	\providecommand{\url}[1]{#1}
	\csname url@rmstyle\endcsname
	\providecommand{\newblock}{\relax}
	\providecommand{\bibinfo}[2]{#2}
	\providecommand\BIBentrySTDinterwordspacing{\spaceskip=0pt\relax}
	\providecommand\BIBentryALTinterwordstretchfactor{4}
	\providecommand\BIBentryALTinterwordspacing{\spaceskip=\fontdimen2\font plus
		\BIBentryALTinterwordstretchfactor\fontdimen3\font minus
		\fontdimen4\font\relax}
	\providecommand\BIBforeignlanguage[2]{{%
			\expandafter\ifx\csname l@#1\endcsname\relax
			\typeout{** WARNING: IEEEtran.bst: No hyphenation pattern has been}%
			\typeout{** loaded for the language `#1'. Using the pattern for}%
			\typeout{** the default language instead.}%
			\else
			\language=\csname l@#1\endcsname
			\fi
			#2}}
	
	\bibitem{WU_2019}
	Y.~Wu, ``Coordinated path planning for an unmanned aerial-aquatic vehicle
	({UAAV}) and an autonomous underwater vehicle ({AUV}) in an underwater target
	strike mission,'' \emph{Ocean Engineering}, vol. 182, pp. 162 -- 173, 2019.
	
	\bibitem{Wolff_2016}
	M.~{Wolff}, R.~T. {Collins}, and Y.~{Liu}, ``Regularity-driven building facade
	matching between aerial and street views,'' in \emph{IEEE CVPR}, June 2016,
	pp. 1591--1600.
	
	\bibitem{Hu_2018_CVPR}
	S.~Hu, M.~Feng, R.~M.~H. Nguyen, and G.~Hee~Lee, ``{CVM-Net}: Cross-view
	matching network for image-based ground-to-aerial geo-localization,'' in
	\emph{IEEE CVPR}, 2018, pp. 7258--7267.
	
	\bibitem{Santos19}
	M.~M. {Dos Santos}, G.~G. {De Giacomo}, P.~L.~J. {Drews-Jr}, and S.~S.~C.
	{Botelho}, ``Satellite and underwater sonar image matching using deep
	learning,'' in \emph{2019 Latin American Robotics Symposium (LARS), 2019
		Brazilian Symposium on Robotics (SBR) and 2019 Workshop on Robotics in
		Education (WRE)}, 2019, pp. 109--114.
	
	\bibitem{Machado_2020}
	M.~{Machado Dos Santos}, G.~G. {De Giacomo}, P.~L.~J. {Drews-Jr}, and S.~S.~C.
	{Botelho}, ``Matching color aerial images and underwater sonar images using
	deep learning for underwater localization,'' \emph{IEEE Robotics and
		Automation Letters}, vol.~5, no.~4, pp. 6365--6370, Oct 2020.
	
	\bibitem{Giacomo_2020}
	G.~G. De~Giacomo, M.~M. dos Santos, P.~L. Drews-Jr, and S.~S. Botelho,
	``Cooperative training of triplet networks for cross-domain matching,'' in
	\emph{2020 Latin American Robotics Symposium (LARS), 2020 Brazilian Symposium
		on Robotics (SBR) and 2020 Workshop on Robotics in Education (WRE)}.\hskip
	1em plus 0.5em minus 0.4em\relax IEEE, 2020, pp. 1--6.
	
	\bibitem{chopra_2005}
	S.~Chopra, R.~Hadsell, and Y.~LeCun, ``Learning a similarity metric
	discriminatively, with application to face verification,'' in \emph{IEEE
		CVPR}, vol.~1, 2005, pp. 539--546.
	
	\bibitem{Goodfellow_2014}
	I.~Goodfellow, J.~Pouget-Abadie, M.~Mirza, B.~Xu, D.~Warde-Farley, S.~Ozair,
	A.~Courville, and Y.~Bengio, ``Generative adversarial nets,'' in
	\emph{Advances in neural information processing systems}, 2014, pp.
	2672--2680.
	
	\bibitem{Hoffer_2015}
	E.~Hoffer and N.~Ailon, ``Deep metric learning using triplet network,'' in
	\emph{International Workshop on Similarity-Based Pattern Recognition}.\hskip
	1em plus 0.5em minus 0.4em\relax Springer, 2015, pp. 84--92.
	
	\bibitem{Thrun_2006}
	S.~Thrun, ``Probabilistic robotics,'' \emph{Communications of the ACM},
	vol.~45, no.~3, pp. 52--57, 2002.
	
	\bibitem{Drews_2014}
	P.~L.~J. {Drews-Jr}, A.~A. {Neto}, and M.~F.~M. {Campos}, ``Hybrid unmanned
	aerial underwater vehicle: Modeling and simulation,'' in \emph{IEEE/RSJ
		IROS}, Sep. 2014, pp. 4637--4642.
	
	\bibitem{Neto2015}
	A.~A. Neto, L.~A. Mozelli, P.~L.~J. Drews-Jr, and M.~F.~M. Campos, ``Attitude
	control for an hybrid unmanned aerial underwater vehicle: A robust switched
	strategy with global stability,'' in \emph{IEEE ICRA}, 2015, pp. 395--400.
	
	\bibitem{mercado2019}
	D.~Mercado, M.~Maia, and F.~J. Diez, ``Aerial-underwater systems, a new
	paradigm in unmanned vehicles,'' \emph{JINT}, vol.~95, no.~1, pp. 229--238,
	2019.
	
	\bibitem{Horn_2020}
	A.~C. {Horn}, P.~M. {Pinheiro}, R.~B. {Grando}, C.~B. {da Silva}, A.~A. {Neto},
	and P.~L.~J. {Drews-Jr}, ``A novel concept for hybrid unmanned aerial
	underwater vehicles focused on aquatic performance,'' in \emph{2020 Latin
		American Robotics Symposium (LARS), 2020 Brazilian Symposium on Robotics
		(SBR) and 2020 Workshop on Robotics in Education (WRE)}, 2020, pp. 1--6.
	
	\bibitem{GAO_2018}
	X.~Gao, S.~Shen, Z.~Hu, and Z.~Wang, ``Ground and aerial meta-data integration
	for localization and reconstruction: A review,'' \emph{Pattern Recognition
		Letters}, vol. 127, pp. 202--214, 2018.
	
	\bibitem{Bansal_2016}
	M.~Bansal, K.~Daniilidis, and H.~Sawhney, ``Ultrawide baseline facade matching
	for geo-localization,'' in \emph{Large-Scale Visual Geo-Localization}.\hskip
	1em plus 0.5em minus 0.4em\relax Springer, 2016, pp. 77--98.
	
	\bibitem{Li_2014}
	A.~Li, V.~I. Morariu, and L.~S. Davis, ``Planar structure matching under
	projective uncertainty for geolocation,'' in \emph{ECCV}, 2014, pp. 265--280.
	
	\bibitem{Majdik_2013}
	A.~L. {Majdik}, Y.~{Albers-Schoenberg}, and D.~{Scaramuzza}, ``{MAV} urban
	localization from google street view data,'' in \emph{IEEE/RSJ IROS}, 2013,
	pp. 3979--3986.
	
	\bibitem{Vis_2014}
	A.~{Viswanathan}, B.~R. {Pires}, and D.~{Huber}, ``Vision based robot
	localization by ground to satellite matching in gps-denied situations,'' in
	\emph{2014 IEEE/RSJ IROS}, Sep. 2014, pp. 192--198.
	
	\bibitem{Lin_2013_CVPR}
	T.-Y. Lin, S.~Belongie, and J.~Hays, ``Cross-view image geolocalization,'' in
	\emph{IEEE CVPR}, 2013, pp. 891--898.
	
	\bibitem{Castaldo_2015}
	F.~Castaldo, A.~Zamir, R.~Angst, F.~Palmieri, and S.~Savarese, ``Semantic
	cross-view matching,'' in \emph{IEEE ICCVw}, 2015, pp. 9--17.
	
	\bibitem{Lin_2015_CVPR}
	T.-Y. Lin, Y.~Cui, S.~Belongie, and J.~Hays, ``Learning deep representations
	for ground-to-aerial geolocalization,'' in \emph{The IEEE Conference on
		Computer Vision and Pattern Recognition (CVPR)}, June 2015.
	
	\bibitem{Workman_2015_CVPR}
	S.~Workman and N.~Jacobs, ``On the location dependence of convolutional neural
	network features,'' in \emph{IEEE CVPRw}, 2015, pp. 70--78.
	
	\bibitem{Workman_2015_ICCV}
	S.~Workman, R.~Souvenir, and N.~Jacobs, ``Wide-area image geolocalization with
	aerial reference imagery,'' in \emph{The IEEE International Conference on
		Computer Vision (ICCV)}, December 2015.
	
	\bibitem{Tian_2017_CVPR}
	Y.~Tian, C.~Chen, and M.~Shah, ``Cross-view image matching for geo-localization
	in urban environments,'' in \emph{IEEE CVPR}, 2017, pp. 3608--3616.
	
	\bibitem{Karkus_2018_ParticleNet}
	P.~Karkus, D.~Hsu, and W.~S. Lee, ``Particle filter networks with application
	to visual localization,'' in \emph{Conference on robot learning}.\hskip 1em
	plus 0.5em minus 0.4em\relax PMLR, 2018, pp. 169--178.
	
	\bibitem{Giacomo_2021}
	G.~G. De~Giacomo, M.~M. dos Santos, P.~L. Drews-Jr, and S.~S. Botelho, ``Guided
	sonar-to-satellite translation.'' \emph{J. Intell. Robotic Syst.}, vol. 101,
	no.~3, p.~46, 2021.
	
	\bibitem{ros}
	\BIBentryALTinterwordspacing
	{Stanford Artificial Intelligence Laboratory et al.}, ``Robotic operating
	system.'' [Online]. Available: \url{https://www.ros.org}
	\BIBentrySTDinterwordspacing
	
	\bibitem{Machado_2019}
	M.~M. dos Santos, G.~G. De~Giacomo, P.~L.~J. Drews, and S.~S. Botelho,
	``Semantic segmentation of static and dynamic structures of marina satellite
	images using deep learning,'' in \emph{2019 8th Brazilian Conference on
		Intelligent Systems (BRACIS)}, 2019, pp. 711--716.
	
	\bibitem{Quigley_2009}
	M.~Quigley, K.~Conley, B.~Gerkey, J.~Faust, T.~Foote, J.~Leibs, R.~Wheeler, and
	A.~Y. Ng, ``Ros: an open-source robot operating system,'' in \emph{ICRA
		workshop on open source software}, vol.~3, no. 3.2.\hskip 1em plus 0.5em
	minus 0.4em\relax Kobe, Japan, 2009, p.~5.
	
	\bibitem{Santos_2019}
	\BIBentryALTinterwordspacing
	M.~M. Santos, G.~B. Zaffari, P.~O. C.~S. Ribeiro, P.~L.~J. Drews-Jr, and
	S.~S.~C. Botelho, ``Underwater place recognition using forward-looking sonar
	images: A topological approach,'' \emph{Journal of Field Robotics}, vol.~36,
	no.~2, pp. 355--369, 2019. [Online]. Available:
	\url{https://onlinelibrary.wiley.com/doi/abs/10.1002/rob.21822}
	\BIBentrySTDinterwordspacing
	
\end{thebibliography}

\end{document}